\documentclass[letterpaper, 10 pt, conference]{ieeeconf}  

\IEEEoverridecommandlockouts                              

\usepackage{graphicx} 
\usepackage{xcolor,soul}
\usepackage{comment}
\usepackage{caption}
\usepackage{subcaption}
\usepackage{amsmath}
\usepackage{url}
\usepackage{fancyhdr}

\usepackage{booktabs}
\usepackage{multirow}

\def\baselinestretch{0.998}

\usepackage{siunitx}
\newcommand{\peter}[1]{\textcolor{blue}{}}
\newcommand{\addresscomments}[1]{#1}

\title{\LARGE \bf
What data do we need for training an AV motion planner?
}

\author{Long Chen$^*$, Lukas Platinsky$^*$, Stefanie Speichert$^*$, Błażej Osiński, \\Oliver Scheel, Yawei Ye, Hugo Grimmett, Luca del Pero, and Peter Ondruska

\thanks{$^*$Equal contribution}

\thanks{Authors are with Lyft Level 5 self-driving division.}
\thanks{Stefanie Speichert is also affiliated with the University of Edinburgh.}
\thanks{Project website: https://planning.l5kit.org}
}

\begin{document}

\renewcommand{\headrulewidth}{0pt}
\fancyhf{}
\fancyhead[C]{Published at 2021 International Conference on Robotics and Automation (ICRA2021)}

\maketitle
\thispagestyle{fancy}
\pagestyle{fancy}

\begin{abstract}
We investigate what grade of sensor data is required for training an imitation-learning-based AV planner on human expert demonstration. Machine-learned planners \cite{bansal2018chauffeurnet} are very hungry for training data, which is usually collected using vehicles equipped with the same sensors used for autonomous operation \cite{bansal2018chauffeurnet}. This is costly and non-scalable. If cheaper sensors could be used for collection instead, data availability would go up, which is crucial in a field where data volume requirements are large and availability is small. We present experiments using up to 1000 hours worth of expert demonstration and find that training with 10x lower-quality data outperforms 1x AV-grade data in terms of planner performance (see Fig.~\ref{fig:intro}). The important implication of this is that cheaper sensors can indeed be used. This serves to improve data access and democratize the field of imitation-based motion planning. Alongside this, we perform a sensitivity analysis of planner performance as a function of perception range, field-of-view, accuracy, and data volume, and reason about why lower-quality data still provide good planning results.

\end{abstract}

\section{INTRODUCTION}

While human drivers can skillfully navigate complex and varied scenarios involving multiple traffic agents safely, motion planning remains one of the hardest problems in autonomous driving. This has motivated the recent interest in planning approaches for Autonomous Vehicles (AVs) based on imitation-learning~\cite{bansal2018chauffeurnet,BojarskiTDFFGJM16}. These methods learn to mimic human behaviour from real-world human driving examples.

These approaches are extremely data-hungry, due to the complexity of the traffic scenarios and their variety: the so-called \emph{long tail} of rare events. Some methods attempt to reduce the need for large volumes of human examples by employing data augmentation techniques, or by synthesizing rare cases in simulation~\cite{wong2020testing}. However, just as the advancements in object detection and classification required large, real-world data sets (e.g., ~\cite{imagenet_cvpr09,lin2014microsoft}), we propose that motion planning also requires \emph{a large corpus of real human driving data}. The problem is that these data are not readily available, and capturing it typically requires a fleet of vehicles equipped with expensive AV-grade sensors, the same that the AV needs to operate in autonomous mode (so that the planner can use the same perception system both at train and test time~\cite{bansal2018chauffeurnet}). This significant barrier to entry stifles progress in the field.

We ask ourselves: do we really need a fleet with AV-grade sensors to collect training data for an AV motion planner? If vehicles equipped with commodity sensors were sufficient, data would be more readily available and this research problem could be democratized.

\begin{figure}
    \centering
    \includegraphics[width=\linewidth]{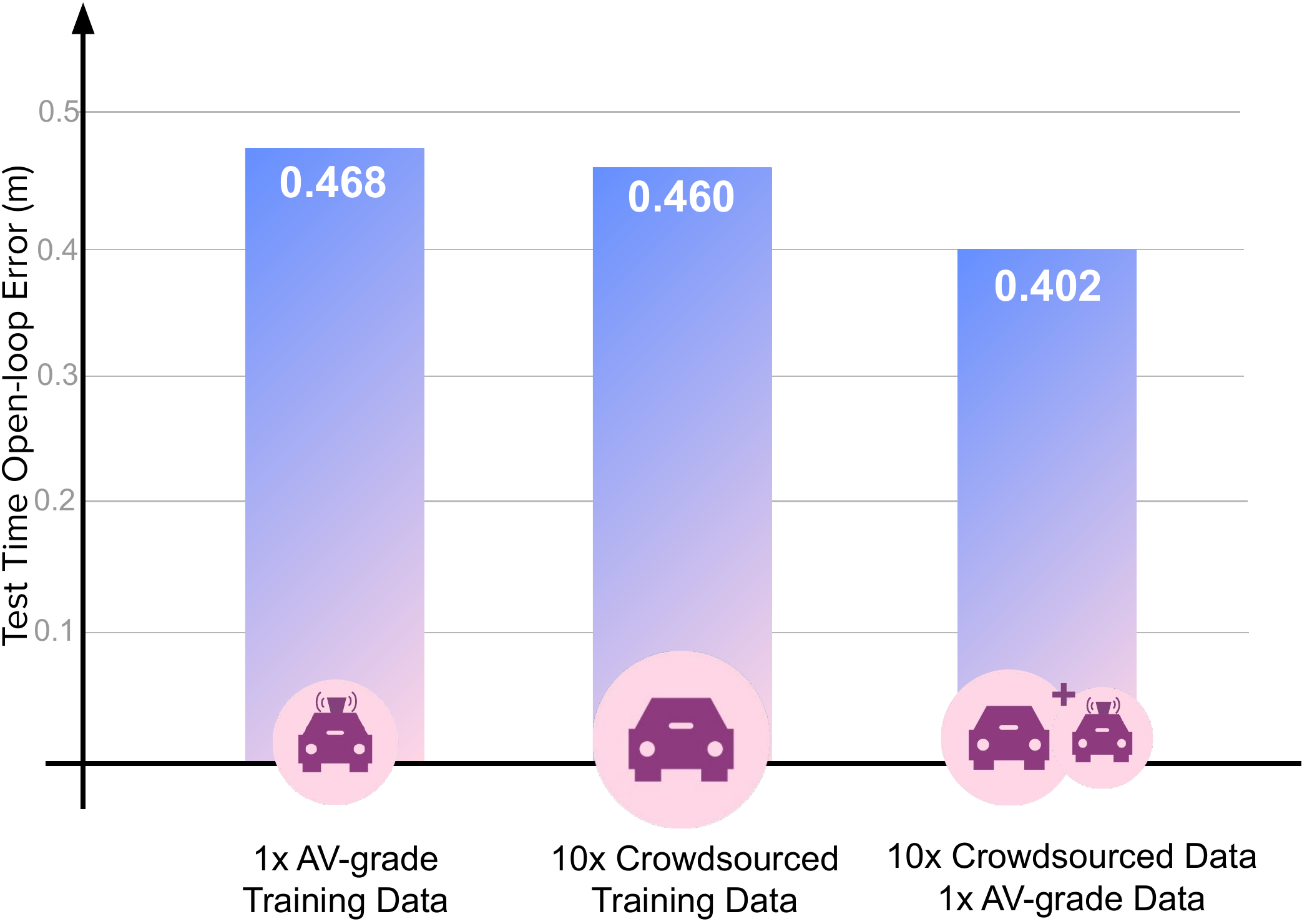}
    \caption{Large data sets of expert driving demonstrations for training motion planners need not come from expensive fleets of AVs. We evaluate motion planner error rate (lower is better), and observe that a planner trained on expert demonstrations collected using AVs (AV-grade, left) has inferior performance to one trained on larger volumes of lower-quality data (middle). Combining the two data sources further improves performance (right), as this bridges the domain gap between the sensors used for training and those used by the AV at test time. This opens the door to crowd-sourced data collection using more affordable sensor configurations than those on AVs.} 
    \label{fig:intro} 
\end{figure}

In this paper, we investigate the data properties we need to train an AV planner for urban operation, by comparing the performance of a state-of-the-art machine-learned planner as we train it on expert demonstration data with varying levels of quality. Modern planning approaches (\cite{bansal2018chauffeurnet, gao_cvpr_2020}) take as input the output of a vehicle's perception system (\emph{perception output}) containing the 3D positions of other traffic agents. The supervision is provided by the human driving the vehicle. By taking AV-grade perception output and progressively limiting its key functional properties, like range and field-of-view (FoV), we simulate the lower-quality data that we could get from a variety of commodity sensors. Our experiments with this data inform what quality we need (and hence which sensors we need), and help us understand the trade-offs between data quantity and quality.

Our key contributions are as follows:
\begin{itemize}
\item We show that training on 10x more data with lower-quality improves on 1x AV-grade data (see Fig.\ref{fig:intro}). This has an important implication: we can collect expert demonstration data for training using vehicles equipped with sensors that are much cheaper than the AV-grade. This opens the door to scalable, cost-effective ways to collect large volumes of human driving examples for teaching AV systems how to drive.

\item We present a sensitivity analysis showing which aspects of data quality are most important for planner performance and why quantity is more important than quality. We derive this from an extensive quantitative evaluation supported by attention-based analysis. 
\end{itemize}

\section{RELATED WORK}


In this paper we build on recent advances in modular machine-learned planning systems.

\textbf{End-to-end vs modular approaches}. Current motion planning methods can be broadly categorized into end-to-end and modular (also known as mid-to-mid).
End-to-end methods consume raw sensor data and output steering commands. A notable example is imitation learning for lane following \cite{BojarskiTDFFGJM16}\addresscomments{, and learning end-to-end driving from simulation \cite{8957584}.}
We refer to \cite{survey_tampuu} for a broad review. Modular methods (e.g. \cite{bansal2018chauffeurnet}) subdivide the problem into sub-tasks (typically, perception, prediction, planning and control), each feeding on the output of the previous one. Sub-tasks are naturally more contained and easier to solve, and having intermediate outputs
makes it easier to interpret the final output. 
More recently, Zeng et al.~\cite{zeng2019end} proposed to bridge the two approaches with an end-to-end architecture producing (and trained on) intermediate outputs. Here, we use modular approach, where the planning module can transfer across different sensor configurations as it takes as input a shared representation, i.e. the output of the perception system.

\textbf{ML planners}.
 An overview of classical planning approaches, including for example expert systems, can be found in \cite{paden2016survey_classical}. Recently, there has been interest in machine-learning (ML) approaches trained on expert demonstration~\cite{bansal2018chauffeurnet, gao_cvpr_2020}, which have the potential to scale with the data.

\textbf{Relationship between planning and perception}.
 Other work has studied the impact of perception quality on planning, either using involved hand-crafted features (e.g. the nuScenes Detection Score in \cite{caesar2020nuscenes}), or as a function of the performance of the planner \cite{philion2020learning}. While the latter work is  relevant, our goal is not to propose a metric to evaluate perception performance, but rather to find what sensors we need to collect training data for planning. Wong et. al~\cite{wong2020testing} simulate perception output for testing the AV planning system. Their focus is on synthesising realistic perception output as if it were captured by AV sensors, and not different levels of perception quality for simulating capturing data from cheaper sensors like we do here.

\textbf{Training with different levels of supervisions}. In our work, we train on a huge amount of lower-quality data and fine-tune on a small amount of high-quality data.
This relates to self-supervised learning, which has recently revolutionized natural language processing \cite{devlin2018bert, brown2020gpt} and is in the process of radically changing computer vision \cite{chen2020simple, he2020moco}. These approaches employ a pre-training phase, in which a neural network is trained on a large collection of unlabeled text/images, and then fine-tuned for the target task on a magnitude smaller collection. Our work is also related to weakly supervised learning, where training happens on data with noisy or incomplete supervision, for example training object detectors on entire images rather than manually annotate 2D bounding boxes~\cite{ren2020arxiv}). In our case, what changes is the quality of the sensors used to collect the training data. 

There are several transfer learning techniques to tackle domain shift between training and testing (e.g.~\cite{5288526}). One option is fine-tuning, which is widely used in literature, e.g.~\cite{vqa, antol2015vqa}. Another is domain adaptation, i.e. aiming to explicitly reduce the domain gap between domains, often used in computer vision domains, for example for synthesizing examples in new domains or style transfer \cite{8099501, hoffman2018cycada, li2018learning}. However, little work has been done to study transfer learning for motion planning.

\textbf{Data availability}. Much previous work relied on proprietary data~\cite{bansal2018chauffeurnet}. Available datasets for planning are few and of moderate size, e.g.~\cite{chang2019argoverse, caesar2020nuscenes}, pointing to the data availability problem mentioned in earlier sections. In our experiments, we use the recent Lyft dataset \cite{houston2020one}, containing 1000 hours of expert demonstrations collected by AVs in urban settings.



\begin{figure*}
    \centering
    \includegraphics[width=\linewidth]{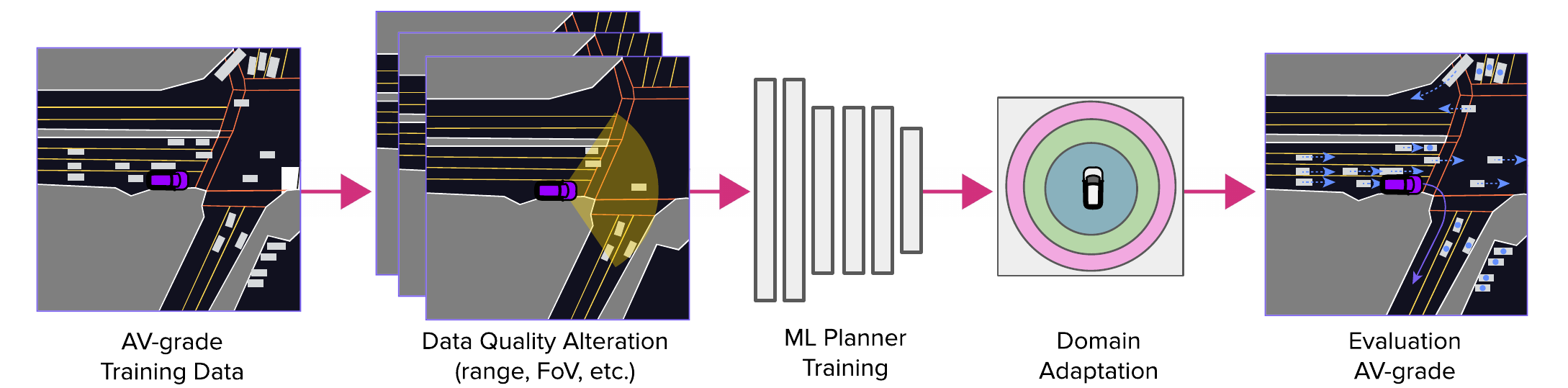}
    \caption{An overview of our methodology (sec.~\ref{method}). Firstly, we collect expert demonstrations using AV-grade sensors, and extract perception information (which contains agent tracks). Secondly, we simulate what this data would look like if collected with lower-quality sensors by reducing the maximum range and field-of-view of the perception information. We then compare the performances of ML planners trained on these training sets with varying quality. This lets us do a comprehensive evaluation without having to build or deploy a wide variety of sensor configurations. To address domain shift, we fine-tune the planner on small amounts of data collected with AV-grade sensors, which is what the planner uses at test time.} 
    \label{fig:method}
\end{figure*}

\section{Methodology}
\label{method}
To understand what type of data we can use to train an AV motion planner, we start by asking two questions: what type of data do we need (\emph{quality}) and how much (\emph{quantity})? 

By data quality we mean the accuracy and robustness of the perception system that the planner uses as input both at test and train time \cite{bansal2018chauffeurnet}. 
The perception sytem outputs the 3D positions of other traffic agents like cars and pedestrians (\textbf{traffic agents}), which provides crucial context, as these positions influence the trajectory followed by the human driver on which the planner is trained.
Quality is driven by the sensors installed on the vehicle, for example, an expensive AV-grade LIDAR can estimate depth much more accurately than a commodity LIDAR, while radars have longer range than a commodity camera system.

To understand data quality requirements, we propose to compare the performance at test time of a state-of-the-art ML-planner as we train it on input data collected with different sensors. However, this would require building and deploying a variety of different sensor configurations, and collecting enough training data with each of them. Instead, we propose to take a dataset of expert driving examples with corresponding AV-grade perception output (Fig.~\ref{fig:method}), and then simulate what this data would look like if it were collected by a wide variety of cheaper sensors. We do this by altering three key dimensions of the perception data:
\begin{enumerate}
\item \textbf{Range}: The maximum distance that sensors can see objects, e.g. 40m, 30m, etc. 
\item \textbf{Field-of-view (FoV)}: How wide sensors can see, e.g. 90°, 180°)
\item \textbf{Geometric accuracy}: How accurate the perception of nearby vehicles is, e.g. the positional and rotational error of detected agents
\end{enumerate}

We therefore start from a dataset of expert-driven examples with AV-grade quality perception, then alter the perception quality along one or more of these three dimensions, and finally train a planner using these data. We repeat this process with varying degrees of quality alteration (see 
Fig.~\ref{fig:method}). A nice property of using the same training samples for all experiments is that it allows us a fair comparison: the difference in performance is only due to difference in quality, and not, for example, due to one dataset containing more diverse examples than the other. 

At test time we always use AV-grade input since, while training data can come from many sources, the goal is to deploy the planner on an actual AV. This introduces domain shift, since the input representation has a different distribution at test time (AV-grade) compared to training (lower-quality). Addressing this is an important part of our methodology. Having investigated data quality with this procedure, we conclude with an analysis on the relationship between data quantity and quality requirements (Sec.~\ref{results}).

\begin{figure}[!b]
    \includegraphics[width=\columnwidth]{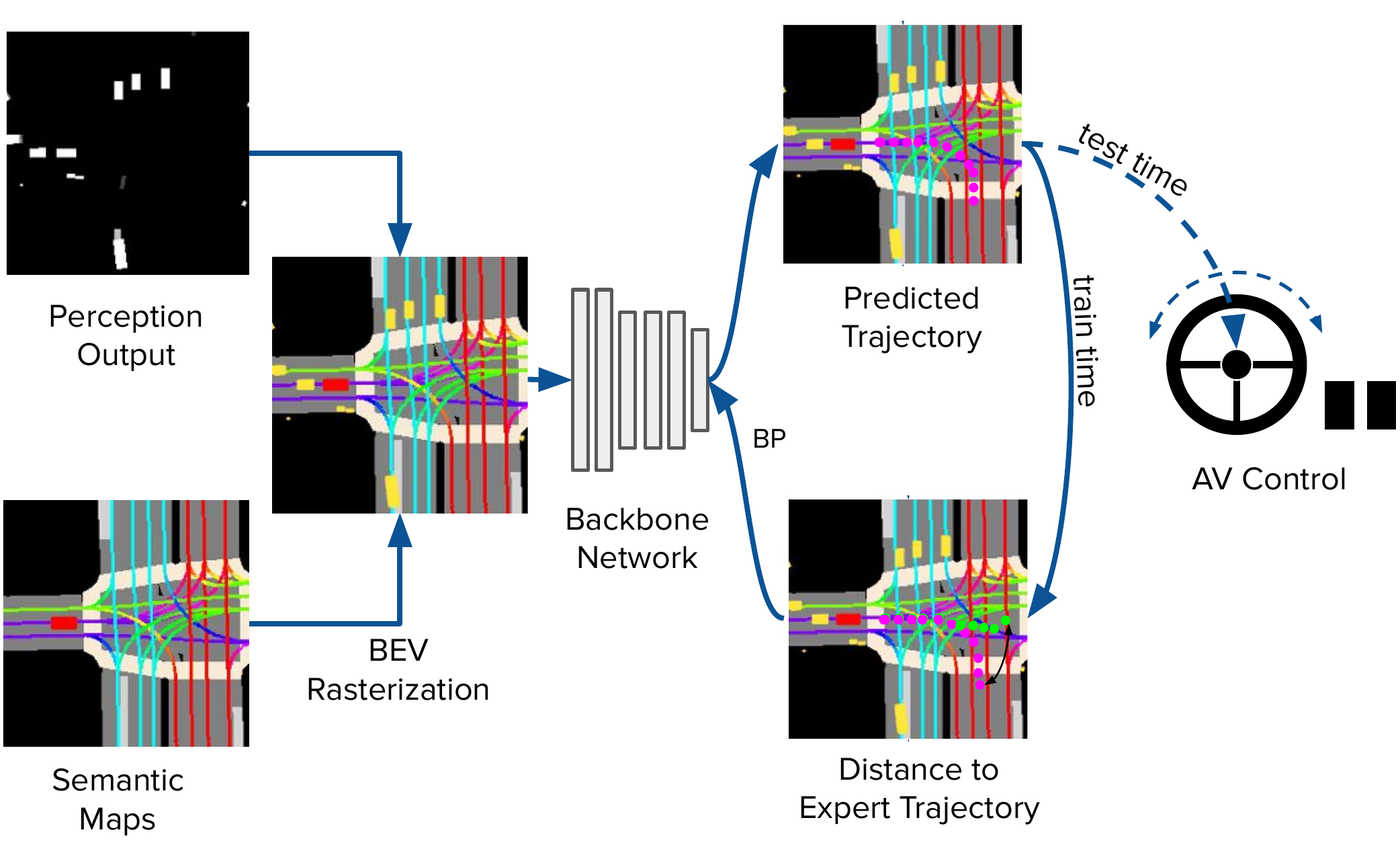}
    \caption{We use a state-of-the-art ML planner (Sec.~\ref{learned_planner}) which takes as input a mid-level representation constructed by rasterizing the agent tracks detected from the perception system on top of an HD semantic map. The network predicts a trajectory for the AV to follow, minimising the distance from the trajectory followed by the expert at training time.} 
    \label{fig:planner} 
\end{figure}

In what follows, we discuss the ML planner we use in our experiments (Sec.~\ref{learned_planner}), how we alter data quality (Sec.~\ref{altering_quality}), and how we address domain shift (Sec.~\ref{domainadaptation}).

\subsection{ML Planner}
\label{learned_planner}
For our experiments we use a state-of-the art ML motion planner trained via imitation learning, similar to e.g. ChaufferNet \cite{bansal2018chauffeurnet}. The input representation to the planner is a birds-eye-view rasterization of the current driving scene centered around the \emph{ego vehicle} (the vehicle carrying the sensors). Agent tracks are rendered as 2D bounding boxes on top of the semantic map of the area, e.g. lane geometry, crosswalks, etc. (see Fig.~\ref{fig:planner}). The network predicts the trajectory to follow for the next $T$ time steps, which we denote by $p = (p_1, \ldots, p_T)$. The training loss minimises the L2 distance between $p$ and the trajectory  $\hat{p}$ followed by the human expert:
\begin{equation}
    l = \sum_{t=1}^{T} |p - \hat{p}|_2
\label{eq:loss}
\end{equation}

We use a ResNet-50 backbone. The network outputs a trajectory for a 1.2 second ($T=12$ steps) prediction horizon. \addresscomments{During training we add synthetic perturbations by alternating the ground-truth trajectories to the left or right side using Ackerman steering \cite{Ackermann} for realistic kinematics}. This was shown to achieve better generalisation, which is needed for closed-loop testing~\cite{bansal2018chauffeurnet}.


\subsection{Altering data quality}
\label{altering_quality}

\textbf{Range and field-of-view}: We reduce the maximum range of the sensors by removing portions of the agent tracks that are beyond a given distance in the AV-grade data (e.g. 20m, 30m, etc.). We reduce the field-of-view (e.g. 90$^{\circ}$, 180$^{\circ}$) by removing agent tracks outside of it (compare Fig.~\ref{fig:method}). The field-of-view is centered along the forward direction of the ego vehicle, e.g. 90$^{\circ}$ degrees denoting 45$^{\circ}$ on each side.

 



\begin{figure*}
    \begin{subfigure}[b]{0.33\textwidth}
      \centering
      \includegraphics[width=0.95\linewidth]{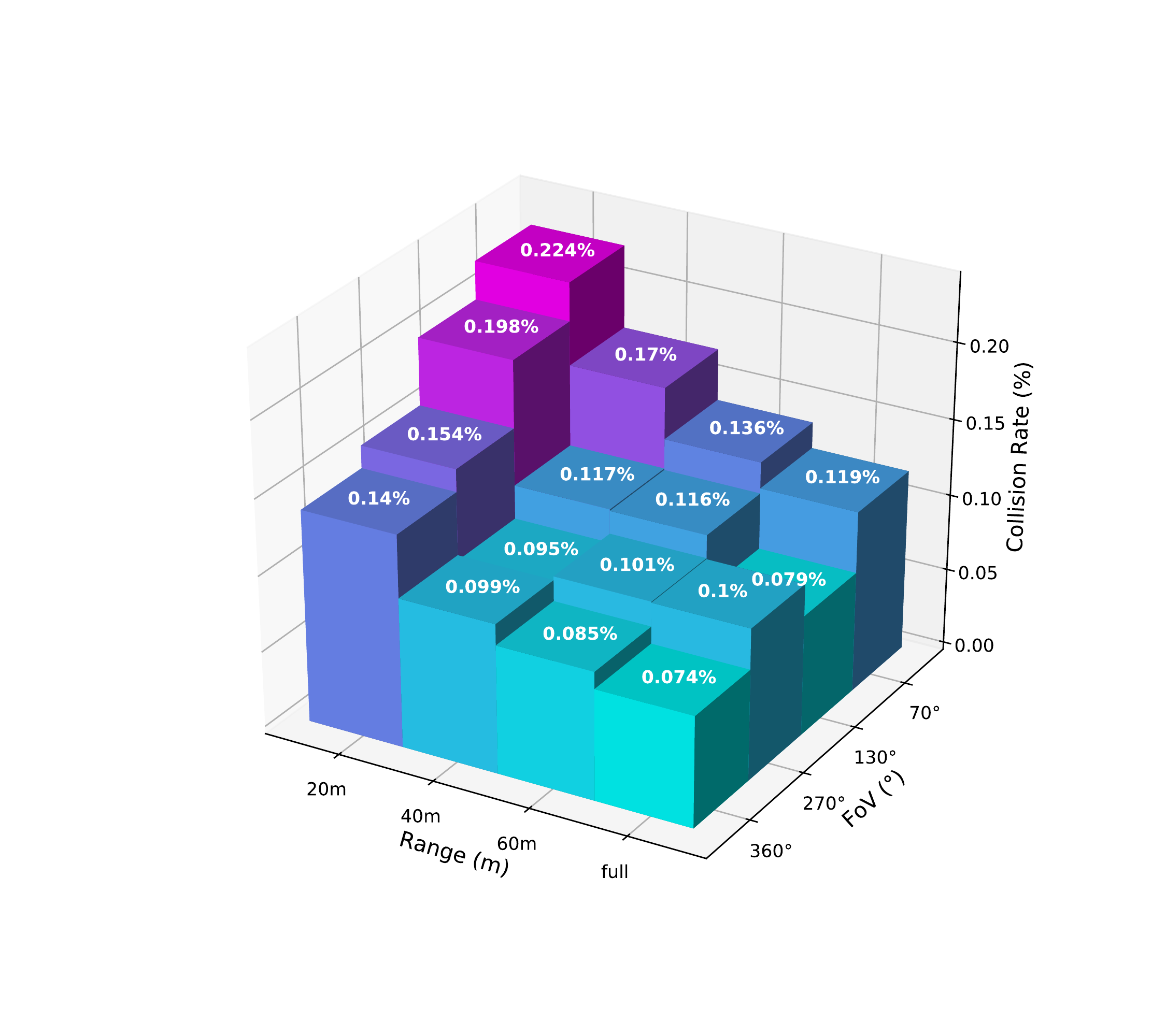}
      \caption{Collision rate}
      \label{fig:quality_close}
    \end{subfigure}
    \begin{subfigure}[b]{0.33\textwidth}
      \centering
      \includegraphics[width=0.95\linewidth]{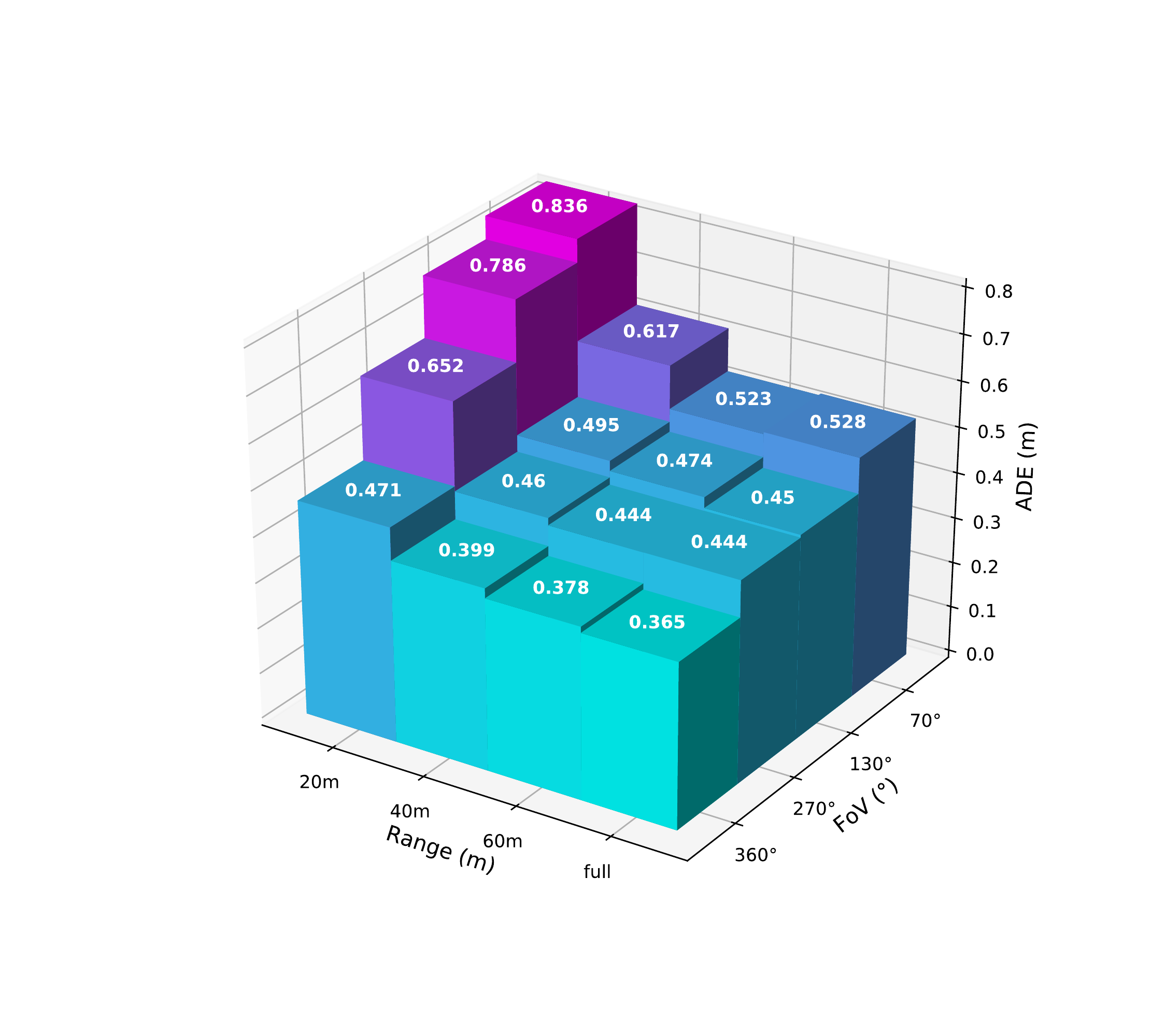}
      \caption{ADE}
      \label{fig:quality_open}
    \end{subfigure}
    \begin{subfigure}[b]{0.33\textwidth}
      \centering
      \includegraphics[width=0.95\linewidth]{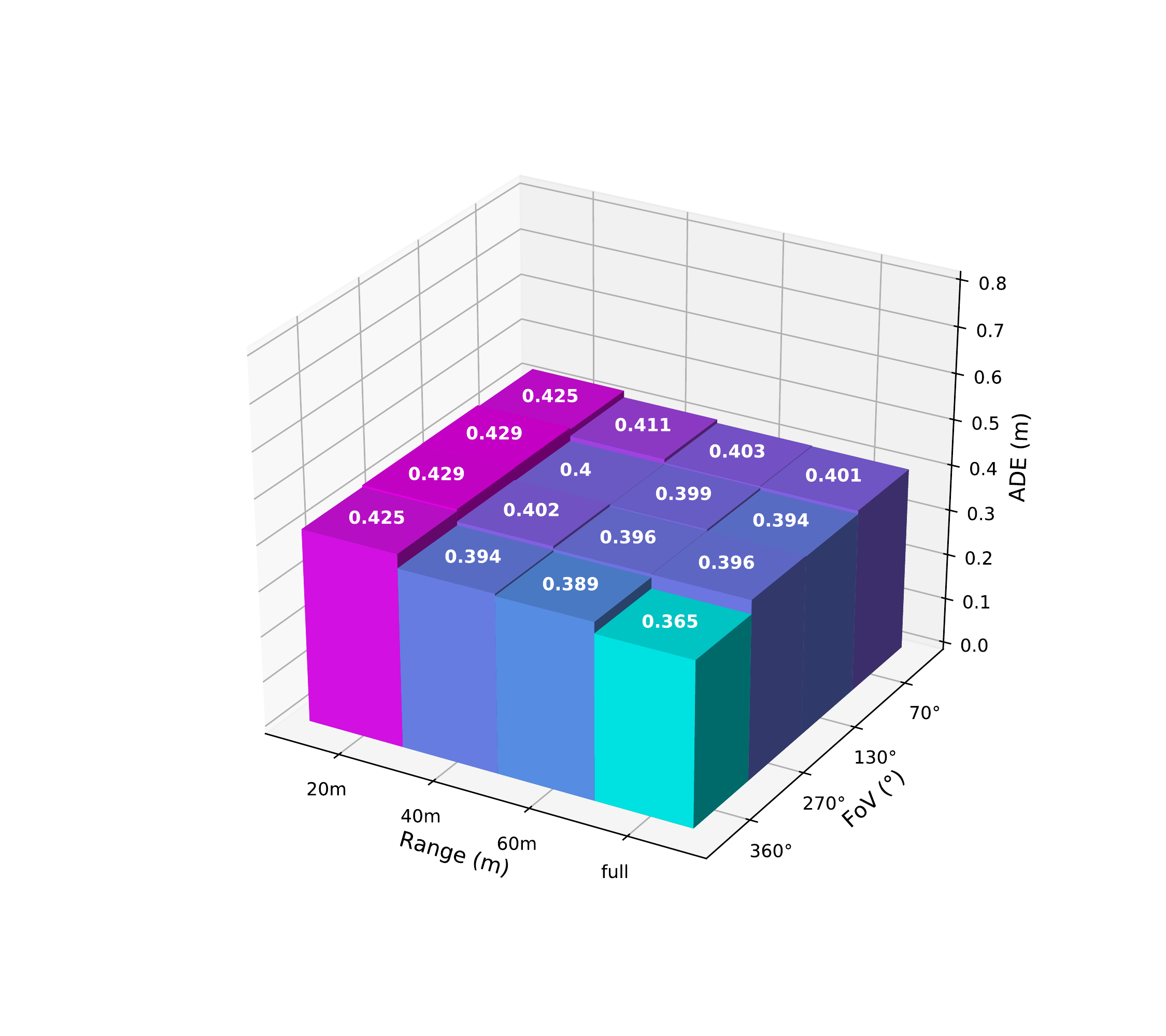}
      \caption{ADE (fine-tuned)}
      \label{fig:after_finetuning}
    \end{subfigure}
    \caption{(a)-(b) Altering range and FoV of the training data (Sec.~\ref{sec:quality_experiments}) impacts performance at test time, with similar trends on both collision rate (a) and ADE (b). There is a significant gap for data with 20m range, while for larger ranges the gap is much smaller. Performance changes smoothly as we alter FoV.
    (c) If we fine-tune (FT) on a small quantity of AV-grade data (10 hours), the gap between training entirely on AV-grade is now much smaller, showing that we can better leverage low quality data if we resolve domain shift (Sec.~\ref{domainadaptation}). Compare this to (b), where we run the same experiment without fine-tuning.
    }
    \label{fig:quality_results}
\end{figure*}

\addresscomments{
\begin{figure}[!b]
\includegraphics[width=8cm]{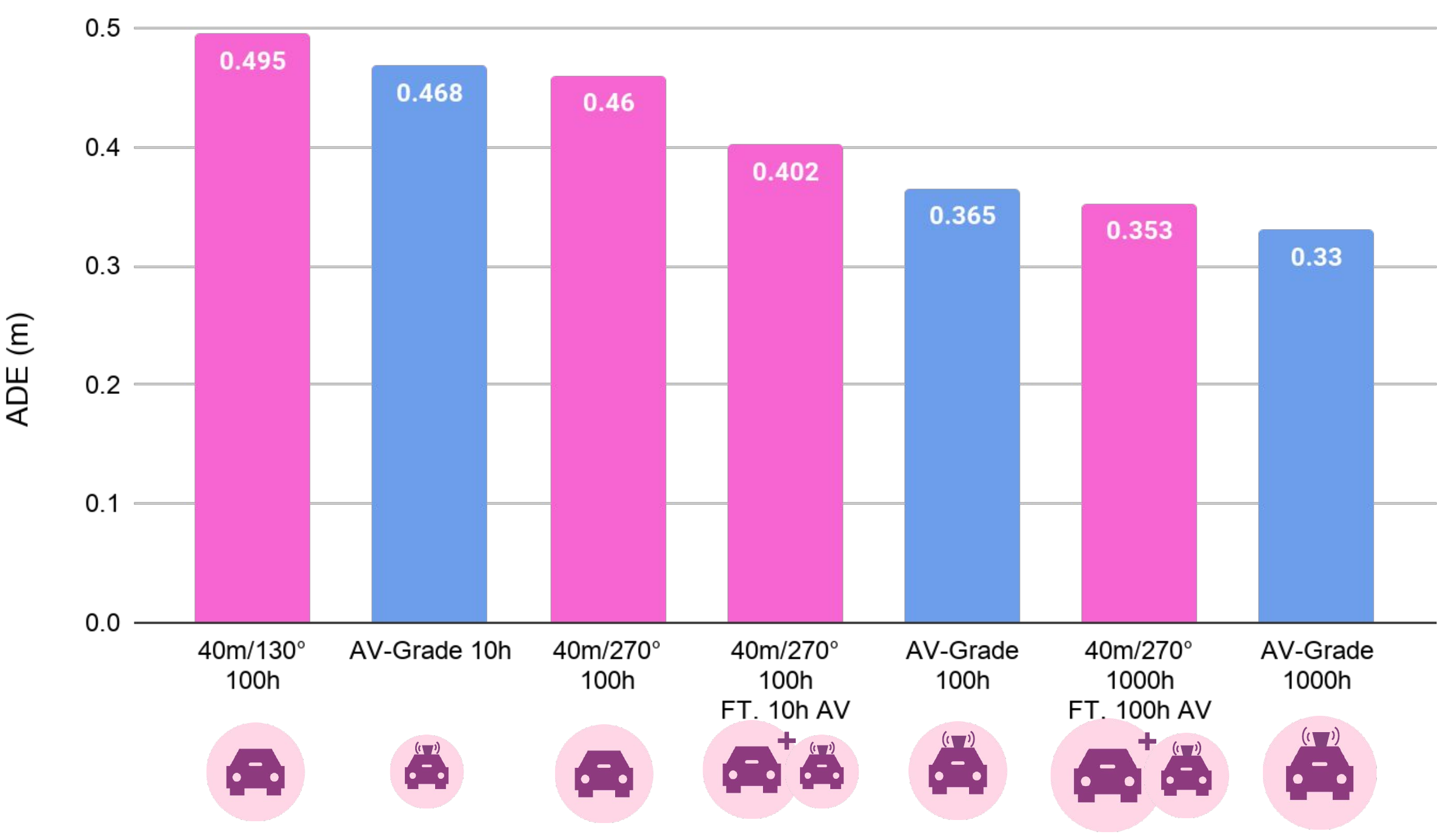}
\caption{Quantity vs quality. Training on 100 hours of low-quality data performs better than training on 10 hours of AV-grade for a wide variety of range/field-of-view combinations. If we fine-tune on 10 hours AV-grade, training on 100 hours of low-quality data approaches 100 hours of AV-grade \addresscomments{(similarly, training on 1000h of low-quality and fine-tuning on 100h of AV-grade approaches 1000h of AV-grade).}} 
\label{close_total_collisions}
\end{figure}
}

\textbf{Geometric accuracy}: We add noise to the position and rotation of the AV-grade perception tracks using noise models approximating the accuracy we would expect from cheaper sensors.

Specifically, we add positional noise to entire agent tracks by applying random offsets to (x,y) agent coordinates in an agent-centered coordinate system. We measure the extent of the positional noise in terms of Intersection over Union (IoU), which is a standard metric for evaluating the accuracy of agent tracks (e.g.~\cite{Geiger2013IJRR}). We specify an IoU noise level that we would like to alter the agent accuracy to. We then add random positional noise such that the noisy position will have the given IoU overlapping with the true position of the agent in the AV-grade data. More specifically, we apply random noise along the longitudinal axis using a uniform distribution, and add noise along the horizontal axis so that the specified IoU level is met. We sample this value once per agent track, and apply it to all agent positions in the track (this is equivalent to shifting the entire track).
This is more realistic than, say, applying i.i.d noise to consecutive agent positions within the same track, as tracks from any sensor are typically processed with some form of smoothing.

For rotational noise, we add a random offset to the angle between the agent direction relative to ego, which we sample from  $r \cdot \mathcal{N}(0,\,1)$, where $r$ is the maximum rotational noise we use in the experiments.

\subsection{Addressing domain shift}
\label{domainadaptation}
Using an input representation with different properties at training and test time introduces domain shift. While our representation is the same in both cases (rasterised BEV, Fig.~\ref{fig:method}), the input agent tracks follow a different distribution and have different levels of noise.

Here, we fine-tune the planner on a relatively small amount of AV-grade data. Intuitively, we first learn how to plan from a large corpus of training data collected with low-quality sensors. Fine-tuning allows us to transfer this wealth of knowledge for use on the AV, which uses different, better sensors at test time. This is an integral part of our proposal of leveraging commodity sensors to collect human driving data. In our experiments we only fine-tune for 2 epochs and use 1/10th of the original base learning rate to avoid overfitting to the small AV-grade dataset.



\begin{figure*}
\begin{subfigure}{0.33\textwidth}
\centering
\includegraphics[width=\linewidth]{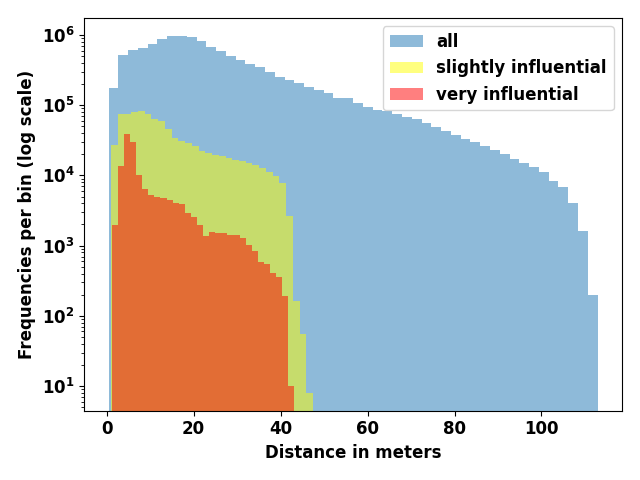}
\caption{}
\label{fig:distance_important}
\end{subfigure}%
~
\begin{subfigure}{0.33\textwidth}
\centering
\includegraphics[width=\linewidth]{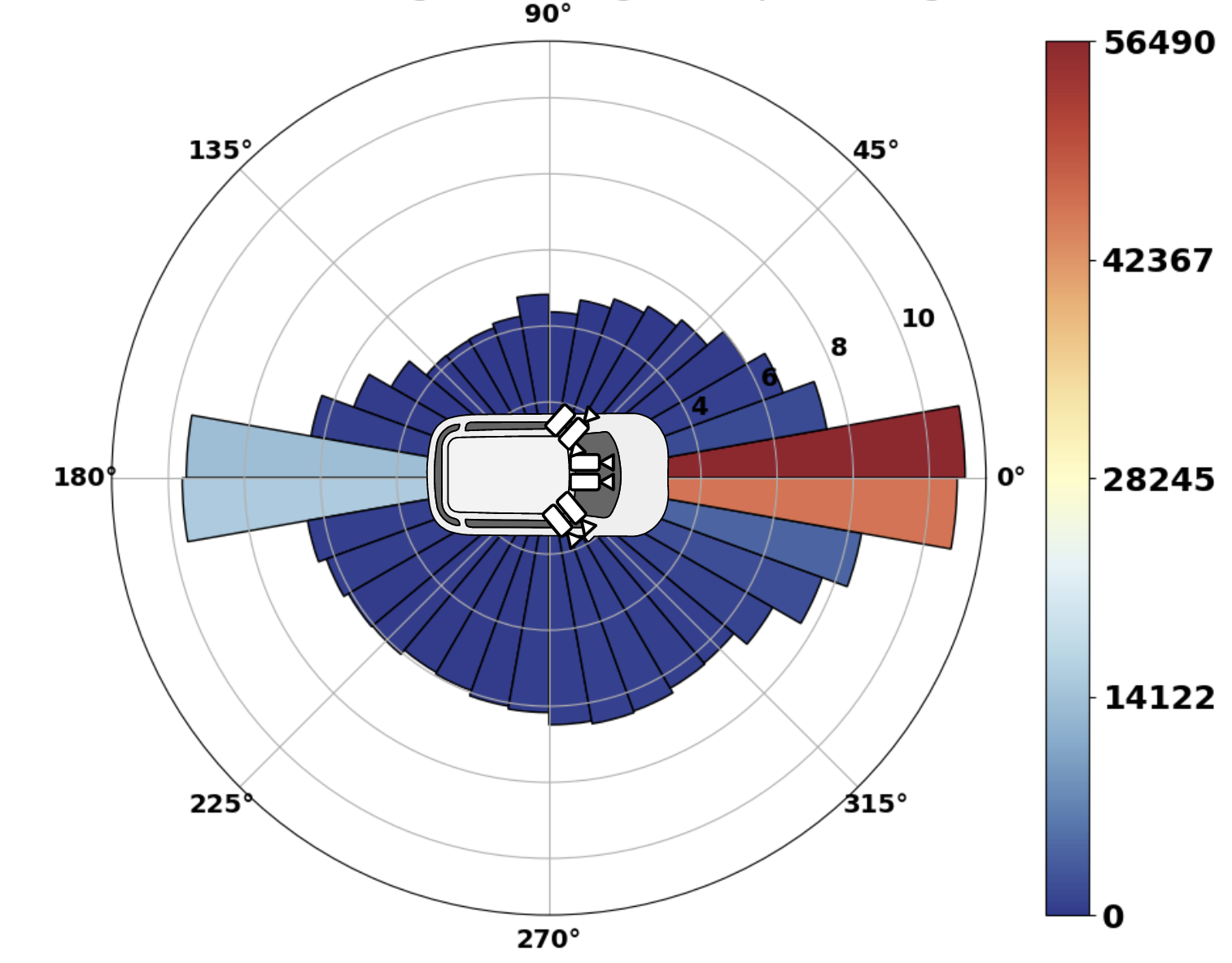}
\caption{}
\label{fig:fov_important}
\end{subfigure}%
~
\vrule
\begin{subfigure}{0.33\textwidth}
\centering
\includegraphics[width=\linewidth]{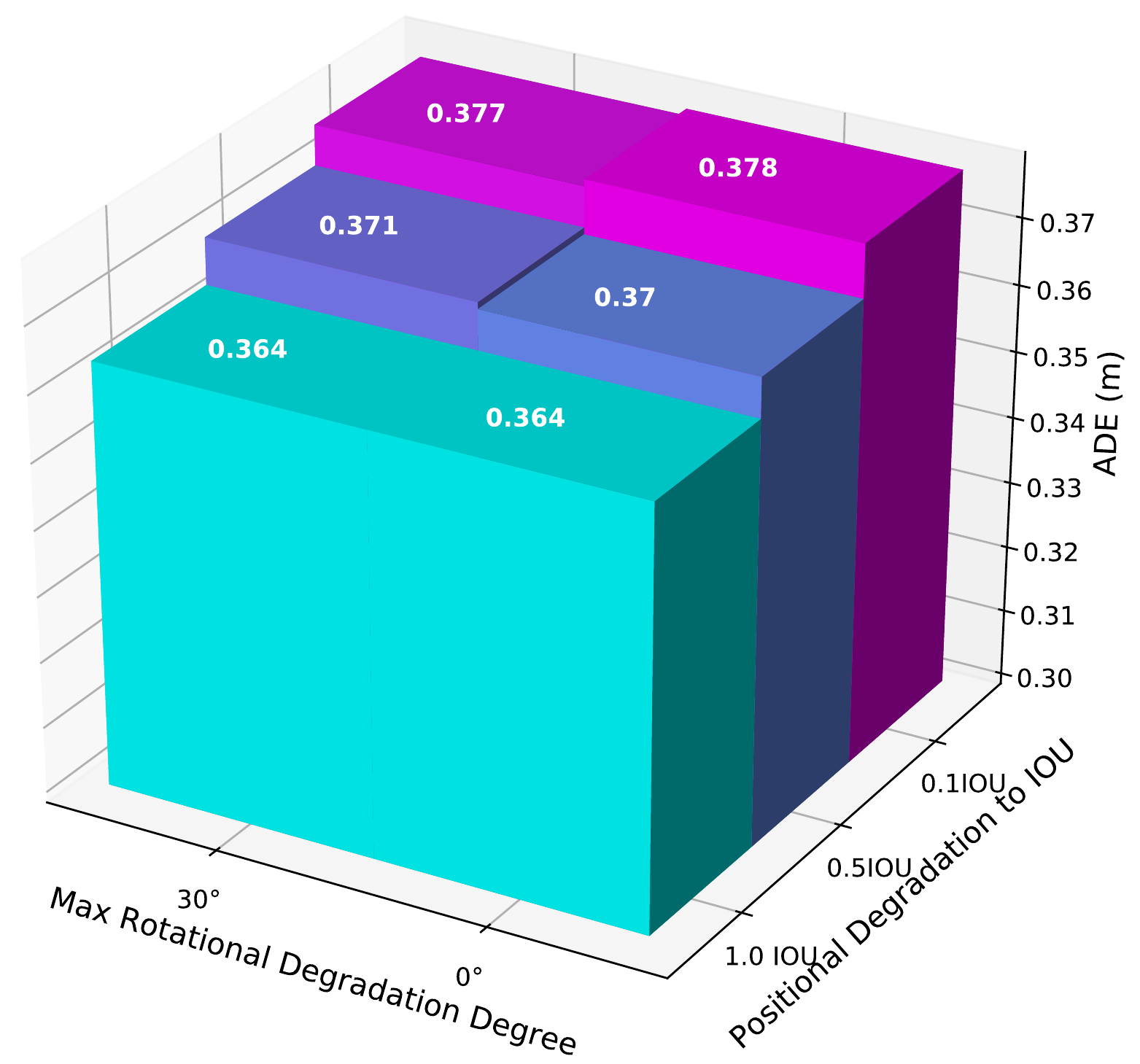}
\caption{}
\label{fig:accuracy_open_loop}
\end{subfigure}
\caption{(a)-(b) The proportion of influential agents (sec.~\ref{metrics}) out of all agents observed in the validation set as a function of their distance to ego (a) and position in the vehicle's field-of-view (b). Most influential agents are within the 40m range and in front of ego (as expected).
(c) ADE as we add positional and rotational error to the agent tracks in the training data (Sec.~\ref{altering_quality}). While test time performance degrades, the impact is negligible compared to altering range and FoV (Fig.~\ref{fig:quality_results}).}
\end{figure*}

\section{RESULTS}
\label{results}
In this section we investigate the impact of training data quality for an ML planner (see Sec.~\ref{sec:quality_experiments}), and the relationship between quantity and quality (see Sec.~\ref{sec:quantity_experiments}). We train our ML planner on either AV-grade data or lower-quality data, and always test on AV-grade data (unless specified otherwise), as our goal is ultimately to deploy the ML planner on a modern AV irrespective of the data it is trained on.
We use standard metrics for this domain (Sec.~\ref{metrics}).

\subsection{Dataset}
\label{sec:dataset}
Our experiments are conducted on the Lyft Level 5 Prediction Dataset\footnote{\url{https://self-driving.lyft.com/level5/data/}} \cite{houston2020one} which contains $>1000$h of expert driving demonstrations with corresponding AV-grade perception output. It was collected in the Palo Alto area by a fleet of 20 AVs, each having 7 cameras, 3 LiDARs, and 5 radars. The data is provided in independent chunks of 25s called \emph{evaluation scenes}, and perception output is refreshed every 10 Hz. In all our experiments, \addresscomments{we use the first 10h, 100h and 1000h data from the L5 training dataset for training, and the original validation dataset for testing}, and let our planner predict at a rate of 10Hz, i.e. every time a new perception observation is available (we call these \emph{steps}).

\subsection{Metrics}
\label{metrics}


\subsubsection{Collision rate (closed-loop evaluation)}
In closed-loop evaluation we allow the ML planner to control the ego vehicle for an extended period of time. Specifically, at time $t=0$ we use the planner to predict the ego location at timesteps $t_1, t_2...t_n$ (sec.~\ref{learned_planner}), and advance the ego to the location at $t_1$. We then feed the planner a new BEV generated at this location to obtain a new prediction. We continue doing this until we reach the end of the evaluation scene, unless the ego collides with an agent or the planner prediction deviates from the ground truth expert trajectory too much where the perception is not reliable anymore. In these last two cases, we stop evaluating the scene. 
The metric we report for closed-loop is the \emph{collision rate}, i.e. the total number of collisions divided by the total number of steps in the dataset.  



\subsubsection{ADE (Open-loop error)}
Unlike in closed loop evaluation, here the planner makes a prediction for each step in a scene independently, and always starts from the ground-truth expert position (no unrolling). We measure the difference between the predicted trajectory and the expert trajectory over the 1.2s prediction horizon using \eqref{eq:loss}, and average over all steps - this is called Average Distance Error (\emph{ADE}).  
Such open-loop error is commonly used in the field (\cite{bansal2018chauffeurnet}, \cite{zeng2019end}).
While it does not provide a holistic view of whether the planner generalises well to the situation when it actually has control over the car like in closed-loop \addresscomments{\cite{10.1007/978-3-030-01267-0_15}}, it is less noisy and much more robust to problems like non-reactivity of the agents and deviations in position impacting perception.


\subsubsection{Agent influence}
Agent influence measures the impact of a specific agent track around the ego vehicle on the output of the planner. It is similar to \cite{philion2020learning} and helps us understand how this influence varies as a function of the relevant data quality dimensions (Sec.~\ref{altering_quality}). For example, the distance of influential agents from the ego vehicle impacts the range we need from the sensors, while their relative position to the ego vehicle impacts FoV. 
We compute influence for agent $a$ as follows: at test time, we first use the planner to predict a trajectory $p$ as usual. We then generate a second raster by masking agent $a$ and let the planner predict a new trajectory $p_a$. We define agent influence to be $\lvert \lvert p-p_a \rvert \rvert_2$, which intuitively indicates how different the planner would behave if it could not see $a$. We further define an agent as \emph{very influential} if this L2 distance is $>$0.1, and \emph{slightly influential} if between 0.1 and 0.01. We compute this for all agents and all steps in the validation dataset.

\begin{figure}[!b]
        \centering
        \begin{subfigure}{0.4\columnwidth}
            \centering
            \includegraphics[width=\columnwidth]{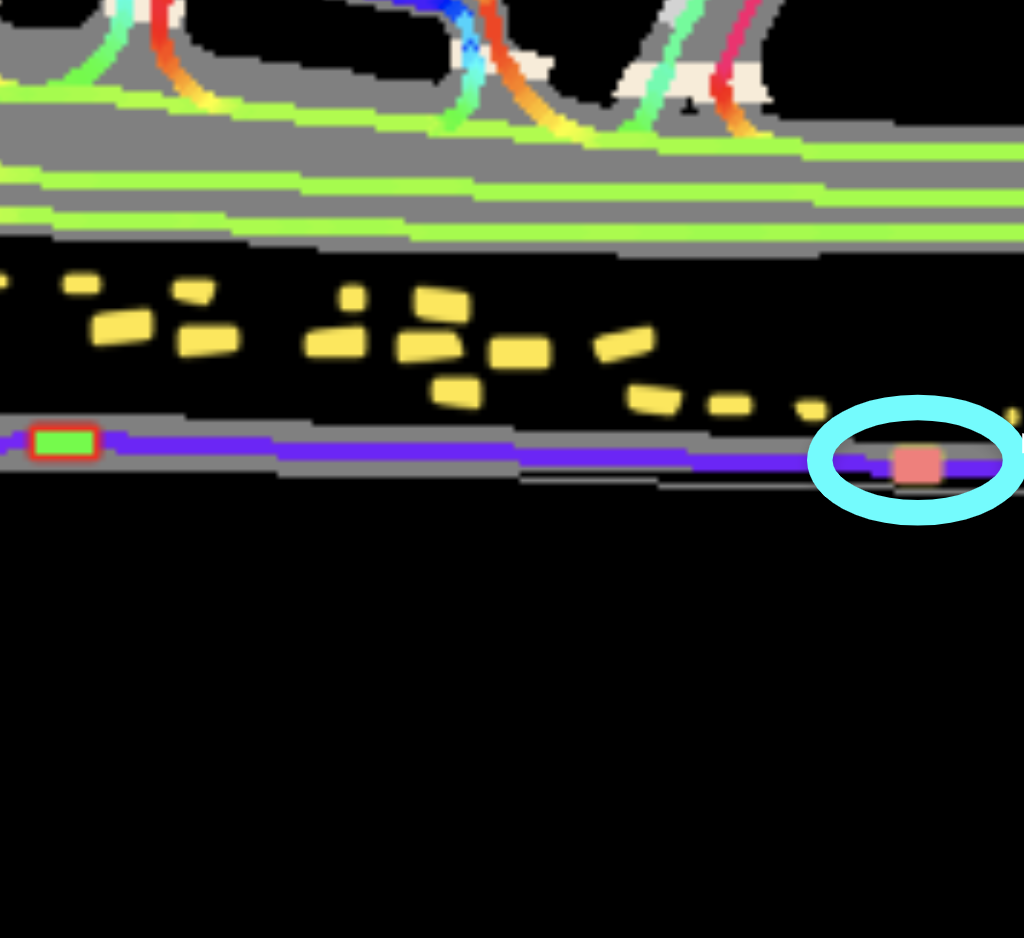}
            \caption[Network2]%
            {}
            \label{fig:distance_all}
        \end{subfigure}
        \begin{subfigure}{0.4\columnwidth}  
            \centering 
            \includegraphics[width=\columnwidth]{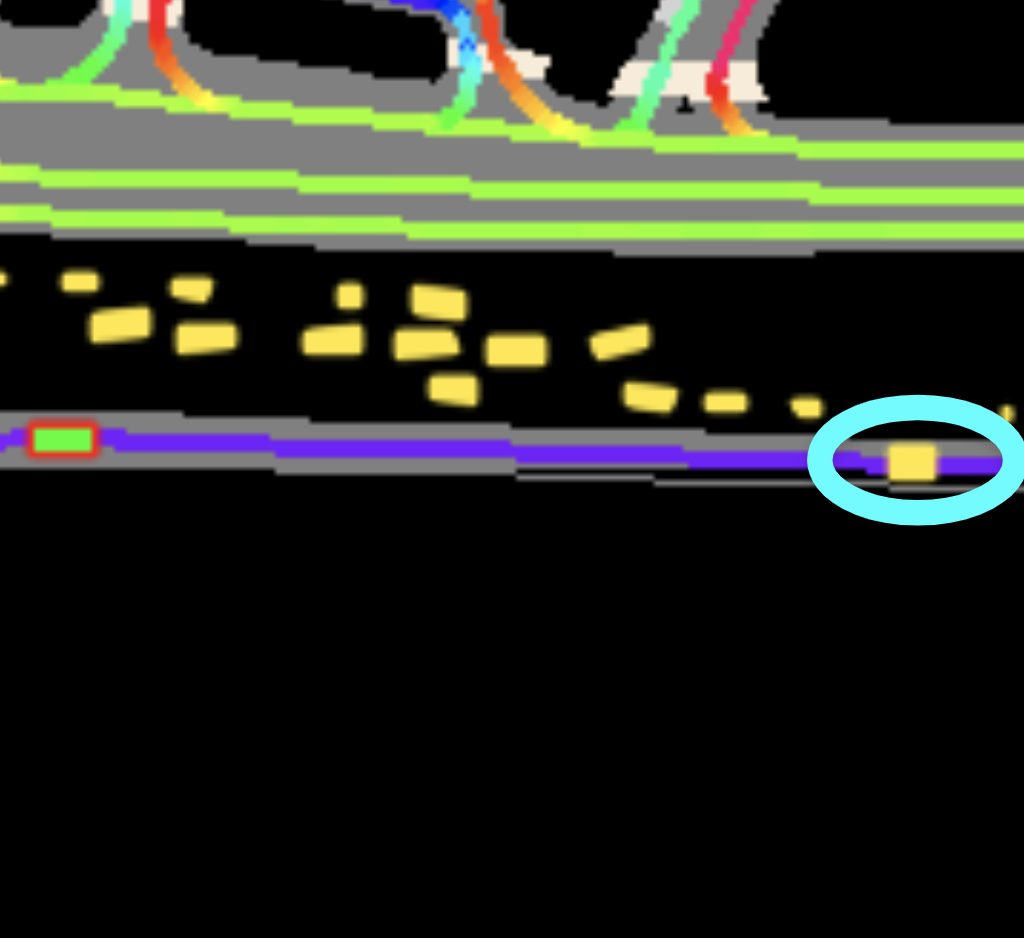}
            \caption[]%
            {}
            \label{fig:distance_small}
        \end{subfigure}
        \begin{subfigure}{0.4\columnwidth}   
            \centering 
            \includegraphics[width=\columnwidth]{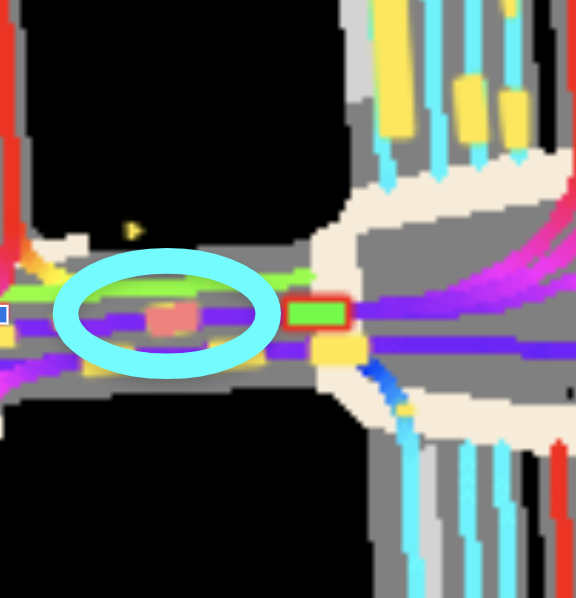}
            \caption[]%
            {}
            \label{fig:mean and std of net34}
        \end{subfigure}
        \begin{subfigure}{0.4\columnwidth}   
            \centering 
            \includegraphics[width=\columnwidth]{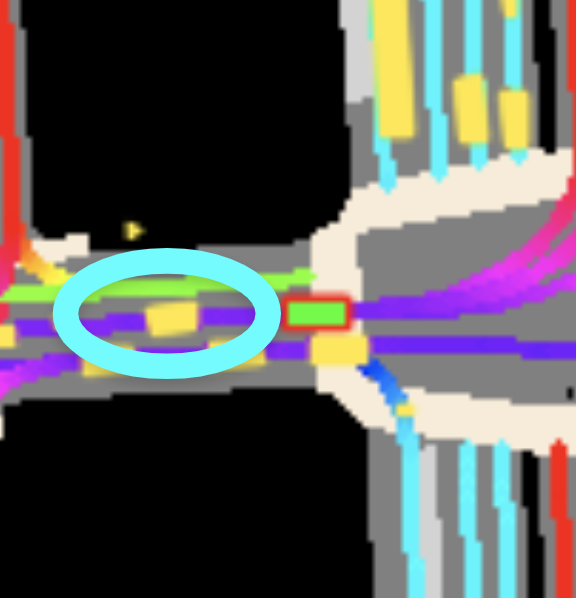}
            \caption[]%
            {}
           \label{fig:mean and std of net35}
        \end{subfigure}
        \caption[]
        {In scenario (a), a model trained on data with AV-grade range is mostly influenced by the only far agent (red) in front of the ego-vehicle (green), while a model trained on 20m range is not (b). We measure this using our agent influence metric (Sec.~\ref{metrics}), i.e. how much the agent influences the planning decision. We observe the same phenomenon for FoV: when training on $360^{\circ}$ (AV-grade), ego is influenced by agents behind (c), but is not when training on $270^{\circ}$ (d).} 
        \label{fig:influence}
\end{figure}

\subsection{Impact of data quality}
\label{sec:quality_experiments}
To measure the impact of data quality, we alter the AV-grade perception tracks in the training data along the three dimensions mentioned in Sec.~\ref{method}. 
For range and FoV (Sec.~\ref{altering_quality}) we use the following buckets:
\begin{itemize}
\item Range: 20m, 40m, 60m, full range (AV-grade)
\item FoV: 70°, 130°, 270°, 360° (AV-grade)
\end{itemize}

We train the planner on all possible permutations of range and FoV (16 in total) and report collision rate and ADE in Fig.~\ref{fig:quality_results}. All results are using the same training set of 100 hours (we always train for 15 epochs), and are evaluated on the full Lyft validation set.  For both metrics, performance increases significantly when the sensor range goes from 20m to 40m, and the FoV from 70° to 130°. However, longer range or wider FoV do not significantly impact performance, and sensors that can achieve 40m range and 130° FoV can already provide valuable training data.

Next, we analyse geometric accuracy (Sec.~\ref{altering_quality}) by using the following buckets:
\begin{itemize}
\item Positional error: 0.1 IOU, 1.0 IOU (zero error)
\item Rotational error: 30°, 0°(zero error)
\end{itemize}
The results in Fig.~\ref{fig:accuracy_open_loop} show that the planner is quite robust to this type of orientation and positional error (1 cm difference in ADE at test time). For this reason, we do not explore geometry accuracy any further in the next experiments. 


Our agent influence metric allows us to gain further insights in these results. When training on AV-grade data, we note that the most influential agents are within the 40m range from the ego vehicle (Fig.~\ref{fig:distance_important}). This is consistent with Fig.~\ref{fig:quality_results}, where training on more than 40m brings negligible improvement. For FoV (Fig.~\ref{fig:fov_important}), agents in front of the ego are the most influential, but we can see a non negligible mass of important agents also at the back. This is aligned with our results in Fig.~\ref{fig:quality_results}, where can see a noticeable improvement in performance when training on $360^{\circ}$ compared to $270^{\circ}$. A qualitative example is shown in Fig.~\ref{fig:influence}.

\subsection{Domain Shift}
We repeat the experiment in Fig.~\ref{fig:quality_open} by fine-tuning each model on 10h of AV-grade data. Results in Fig.~\ref{fig:after_finetuning} show how much we can mitigate domain shift by exploiting a very limited amount of data from the target AV sensor configuration, and even models trained with only 20m range now become much more competitive. 

\subsection{Relationship between data quality and quantity}
\label{sec:quantity_experiments}

In this section we analyse the relationship between data quantity and quality. 
In Fig.~\ref{fig:quality_results} we saw a performance degradation when we alter quality while training on the same amount of data (100h). Fig.~\ref{close_total_collisions} clearly shows that training on 1000 hours low-quality data (10x) performs better at test time than 100 hours AV-grade (1x) for a variety of range/FoV configurations. As we fine-tune on additional 100h AV-grade data (Sec.~\ref{domainadaptation}), we see that performance approaches training on 1000h AV-grade.

\section{Conclusion}
We have shown that it is possible to train an AV planner on human expert demonstrations collected with sensors that are of lower quality than AV sensors. Our results suggest that data quantity trumps quality, and that it is better to have a lot of expert demonstrations with lower-quality, than a smaller amount of demonstrations that are AV-grade. For example, 100 hours with 40m range and $270^{\circ}$ field-of-view data outperforms 10 hours of AV-grade quality (Fig.~\ref{close_total_collisions}). Moreover, it approaches 100 hours AV-grade (the same amount) after fine-tuning on a small amount of AV-grade data, showing we can tackle domain shift. 
All in all this shows the promise of a crowd-sourcing approach for democratising the collection of training data for training AV motion planners. 

We can use our results to inform the choice of sensors for collecting expert demonstration data, choosing the trade-off between sensor complexity (and cost) and expected performance in urban environments (Fig.~\ref{fig:after_finetuning}). The fact that several sensor configurations are competitive shows that we could combine data collected with a variety of heterogeneous sensors for training. Furthermore, we can apply the methodology presented in this paper to analyse data coming from different environments (e.g. highways) to understand the trade-offs between the sensor configurations in this setting. 

In future work, we plan to experiment with data collected using real commodity sensors, rather than using our data-altering approach. Moreover, we will study how these results generalise to different areas and conditions.

\addtolength{\textheight}{-10cm}   




\vspace{10pt}
\section*{ACKNOWLEDGMENTS}
We would like to acknowledge and thank Sergey Zagoruyko for the valuable suggestions for this paper.



\bibliographystyle{IEEEtran}
\bibliography{references}

\end{document}